\documentclass[10pt]{article}
\usepackage{chngcntr}
\usepackage{amsmath}
\usepackage{amssymb}

\usepackage{graphicx}

\usepackage{cite}

\usepackage{color} 

\usepackage[labelfont=bf,labelsep=period,justification=raggedright]{caption}

\makeatletter
\renewcommand{\@biblabel}[1]{\quad#1.}
\makeatother

\date{}

\usepackage{epstopdf}
\usepackage{wrapfig}
\usepackage{epsfig, subfigure}
\usepackage{color}

\newcommand{\BEQ}{\begin{equation}}
\newcommand{\EEQ}{\end{equation}}
\newcommand{\BEA}{\begin{eqnarray}}
\newcommand{\EEA}{\end{eqnarray}}
\newcommand{\BGA}{\begin{gather}}
\newcommand{\EGA}{\end{gather}}
\newcommand{\comment}[1]{}

\newcommand{\revise}[1]{{\color{black} #1}}
\newcommand{\revisetwo}[1]{{\color{black} #1}}
\newcommand{\revisethree}[1]{{\color{black} #1}}

\begin{document}

\begin{flushleft}
{\Large
\textbf{ \revise{Understanding Confounding Effects in Linguistic Coordination: an Information-Theoretic Approach} }
}
\\
Shuyang Gao$^{1,\ast}$, 
Greg Ver Steeg$^{1}$, 
Aram Galstyan$^{1}$
\\
\bf{1} Information Sciences Institute, University of Southern California, Marina del Rey, CA
\\
$\ast$ E-mail: Corresponding sgao@isi.edu
\end{flushleft}

\section*{Abstract}
We suggest an information-theoretic approach for measuring stylistic coordination in dialogues. The proposed measure has a simple predictive interpretation and can account for various confounding factors through proper conditioning. We revisit some of the previous studies that reported strong signatures of stylistic accommodation, and find that a significant part of the observed coordination can be attributed to a simple confounding effect -  length coordination. Specifically, longer utterances tend to be followed by longer responses, which gives rise to spurious correlations in the other stylistic features. We propose a test to distinguish correlations in length due to contextual factors (topic of conversation, user verbosity, etc.) and turn-by-turn coordination. We also suggest a test to identify whether stylistic coordination persists even after accounting for length coordination and contextual factors. 


\section*{Introduction}
Communication Accommodation Theory~\cite{Giles1991} states that people tend to adapt their communication style (voice, gestures, word choice, etc.) in response to the person with whom they interact.  Originally, experiments on linguistic accommodation were confined to small scale laboratory settings with a handful of participants. 
The recent proliferation of digital (or digitized) communication data offers an opportunity to study nuances of human communication behavior on much larger scales. A number of recent studies have indicated presence of stylistic coordination in communication~\cite{Ireland2010,Gonzales2010,Danescu-Niculescu-Mizil2011WWW,Danescu-Niculescu-Mizil2012WWW}, where one person's use of a linguistic feature (e.g. prepositions) increases the probability that a response will include the same feature. Linguistic style  coordination (or {\em matching})  has been used to predict relationship  stability~\cite{Ireland2010} and negotiation outcomes~\cite{Taylor2008}, understand group cohesiveness~\cite{Gonzales2010}, and infer relative social status and  power relationships among individuals~\cite{Danescu-Niculescu-Mizil2012WWW}.

Most reports of linguistic style coordination have been based on correlational analysis. Thus, such claims are susceptible to various confounding effects. For instance, it is known that there is significant length coordination in dialogues,  in the sense that a longer utterance from user $Y$ tends to solicit a longer response from user $X$~\cite{Niederhoffer2002}. Thus, if the probability of an utterance containing a feature, e.g. \textit{prepositions} or \textit{words whose second letter is ``r''}, depends only on length, this will create the illusion of stylistic coordination on the given feature.

Here we attempt to remedy the problem and propose an information-theoretic framework for characterizing stylistic coordination in dialogues. Namely, given a temporally ordered sequence of utterances (verbal or electronic statements depending on the context) by two individuals, we characterize their stylistic coordination with   {\em time-shifted}  mutual information. The proposed coordination measure characterizes the dependence between the stylistic features of the original post and the response. In addition, we provide a computational framework to account for confounders when measuring  stylistic coordination. 

We revisit some of the case studies where linguistic coordination was reported and demonstrate that a significant part of the observed correlations in linguistic features can indeed be explained by length coordination rather than stylistic accommodation. In particular, most stylistic features that exhibit statistically significant correlation exhibit little to no correlation after length coordination has been taken into account. 

We also focus on the observed length correlations, and  examine whether it is due to turn-by-turn coordination between the participants, or can be attributed to other contextual factors.  We construct a statistical permutation test and demonstrate unequivocally that turn-by-turn length coordination in dialogues indeed takes place. Finally, we develop a similar test for stylistic features, and demonstrate that \revise{at least for one of the datasets}, the remnant coordination (after conditioning on length) cannot be explained by contextual factors alone and has to be due to turn-by-turn level coordination between the speakers.

\section*{Measuring Stylistic Coordination}
\subsection*{Representing Stylistic Features}
\label{RSC}
To represent stylistic features in utterances, we use Linguistic Inquiry Word  Count (LIWC)~\cite{Pennebaker2007}, which is a dictionary-based encoding scheme that has been used extensively  for evaluating emotional and psychological dimensions in various text corpora. The latest version of the LIWC dictionary contains around 4500 words and word stems. Each word or word stem belongs to one or more word categories or subcategories. Various LIWC categories include positive and negative emotion, function words, pronouns, articles, and so on.  Here we focus on eight LIWC categories that have been used in previous studies~\cite{Danescu-Niculescu-Mizil2012WWW}: articles, auxiliary verbs, conjunctions, high-frequency adverbs, impersonal pronouns, personal pronouns, prepositions, and quantifiers. Utterances are represented as eight-component binary vectors indicating the presence or absence of each linguistic marker~\cite{Danescu-Niculescu-Mizil2012WWW}. 

\subsection*{Information-theoretic measure  of coordination} 
\label{IT}
Each dialogue is a sequence of utterance exchanges between two participants. Following ~\cite{Danescu-Niculescu-Mizil2011WWW,Danescu-Niculescu-Mizil2012WWW} we binarize the stylistic features of utterances, so that a dialogue is represented as $\{o_k^m, r_k^m\}_{k=1}^K$, where $o_k^m,r_k^m=\{0,1\}$ indicate the absence or presence of the stylistic marker $m$, and $K$ is the total number of exchanges in a dialogue.  Since we focus on coordination between the same stylistic markers, we will drop the superscript $m$ from now on. We use the convention $O$ to represent the originator -- the person who is producing the original utterance in a single exchange, $R$ to represent the respondent -- the person who is replying to the originator. 



Let $p(o,r)$ be the joint distribution of the random variables $O$ and $R$. We characterize the amount of stylistic coordination using Mutual Information (MI)~\cite{CoverThomas2006}; see \revisethree{S1 File} for a brief overview of basic information theoretic concepts: 
\BEA
I(O:R) = H(O) - H(O|R)
\label{eq_MI}
\EEA
where $H(O)=-\sum p(o)\log p(o)$ is the Shannon entropy of $O$, and $H(O|R)$ is the entropy of $O$ conditioned on $R$. Note that in our case the arguments are temporarily ordered: $O$ is always the initial utterance, and $R$ is the response, so that Eq.~\ref{eq_MI} in fact defines time-shifted mutual information. Thus, even though MI is symmetric with respect to  its argument, the coordination between two users may be asymmetric. 

\revise{ Recall that mutual information between two variables measures the average reduction in the uncertainty of one variable, if we know the other variable. Thus, in essence, the proposed measure of stylistic coordination quantifies how the use of a marker $m$ in an utterance of $O$'s  can help to predict $R$'s usage of $m$ in the immediate response. In contrast to linear correlation measures, mutual information is well suited for handling strongly non-linear dependencies.}


We measure the correlation between two variables \textit{after} conditioning on a third variable, $Z$, via Conditional Mutual Information, defined as 
\BEQ
I(O:R|Z) = H(O|Z) - H(O|R,Z).
\label{eq:cmi}
\EEQ
Below we will use CMI to account for the confounding effect of the utterance length by conditioning on it. Namely, the actual stylistic accommodation, after accounting for the length coordination, is given by $I(O:R|L_R)$, where $L_R$ is the length of the utterance by user $R$.

\subsection*{Estimating mutual information from data}  
\label{Estimation}
Given a set of samples $\{o_k, r_k\}_{k=1}^K$, our goal is to estimate mutual information between $O$ and $R$. We could do this by first calculating the empirical distribution $p(o,r)$ and then using Eq.~\ref{eq_MI}. However, it is known that this naive {\em plug-in} estimator tends to underestimate the entropy of a system. Instead, here we use the statistical bootstrap method introduced by DeDeo et al.~\cite{dedeo}, which attempts to reduce the bias of the naive estimator by estimating a bootstrap correction term. The estimate of bias comes from comparing the entropy of the empirical distribution to estimates of entropy from several bootstrap datasets drawn randomly according to the empirical distribution. See \cite{dedeo} for more details.


While the above discrete estimator works well for evaluating mutual information between discrete stylistic variables, it is not very useful for evaluating mutual information between two length variables, due to limited number of samples we have. Instead, we will use a continuous estimator introduced by Kraskov et al.~\cite{kraskov2004estimating}.  This non-parametric estimator searches the $k$-nearest neighbors to each point, and then average the mutual information estimated from the neighborhood of each point. It has been shown that this estimator is asymptotically unbiased and consistent. 
Discussion of different entropy estimators can be found in~\cite{Wang2009} and references therein.
  

\section*{Length  as a confounding factor}  
\label{result_1}
We  applied our coordination measures to two datasets previously studied in~\cite{Danescu-Niculescu-Mizil2012WWW}: oral transcripts from the Supreme Court hearings, and discussion among Wikipedia editors. In the Supreme Court Data, there are 11 Judges and 311 Lawyers conversing with each other. We obtain 51,498 utterances from all the dialogues among 204 cases. In the Wikipedia dataset, users are classified into two categories, Administrators, or \textit{\em Admins}, and {\em non-Admins}. All of the users interact with each other on Wikipedia \textit{talk} pages, where they discuss issues about specific Wikipedia pages. We focus on dialogues where each participant make at least two exchanges within a dialogue, which results in over 30,000 utterances. 

Ideally, we would like to calculate linguistic accommodation between any pair of individuals $O$ and $R$ who have participated in a dialogue. Unfortunately,  most pair-wise exchanges are rather short and do not produce sufficient samples for evaluating mutual information or conditional mutual information. Instead, we group the individuals according to their roles, and then use aggregated samples to calculate stylistic coordination between the groups.  The groups correspond to Judges and Lawyers for the Supreme Court data, and Admins and non-Admins for the Wikipedia data. 

\begin{figure}[htbp]
\centering
\subfigure[][]{
    \includegraphics[width=.45\columnwidth]{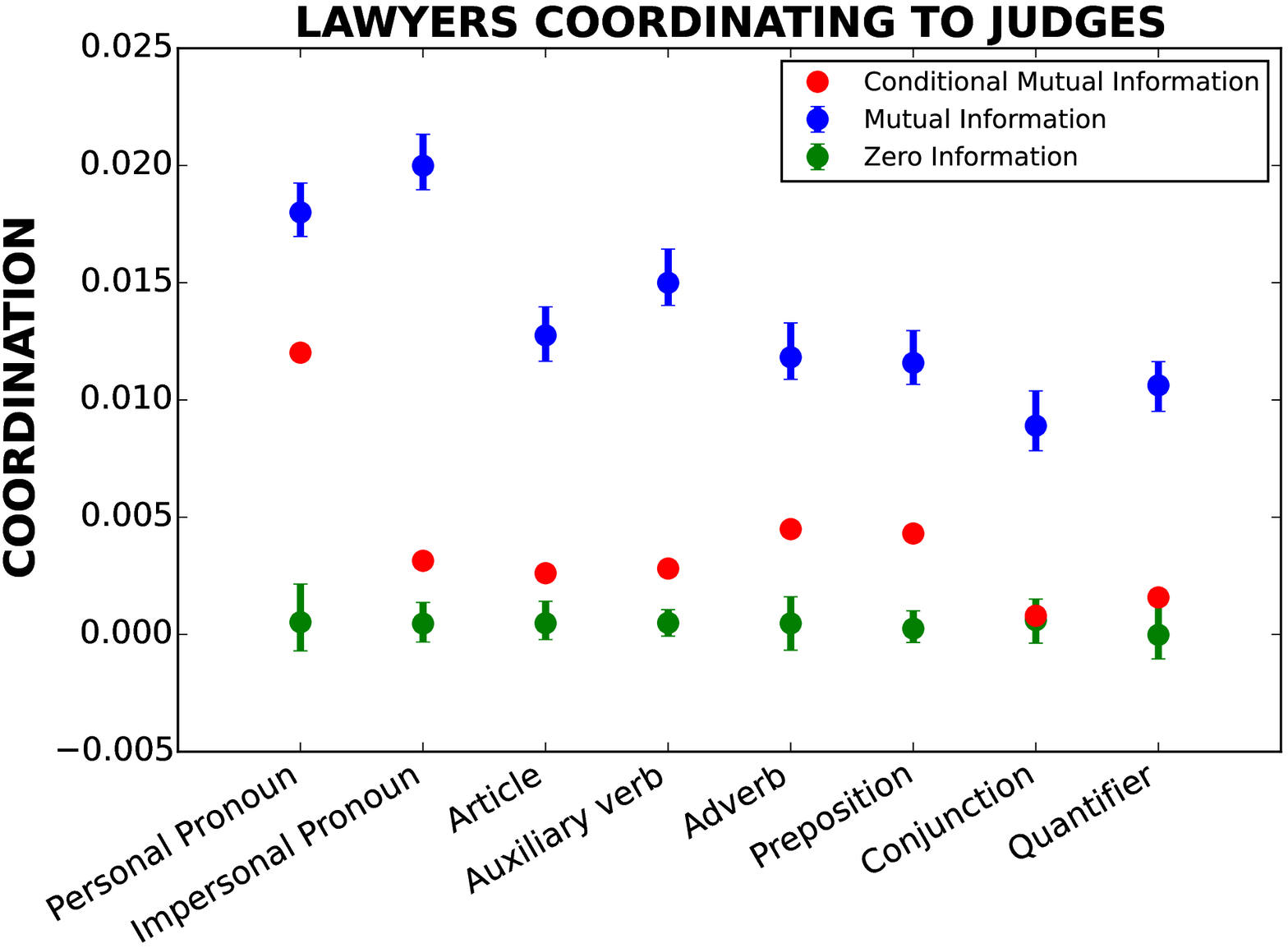}
\label{fig_supreme_court_style_a}
   } 
  \subfigure[][]{%
   \includegraphics[width=.45\columnwidth]{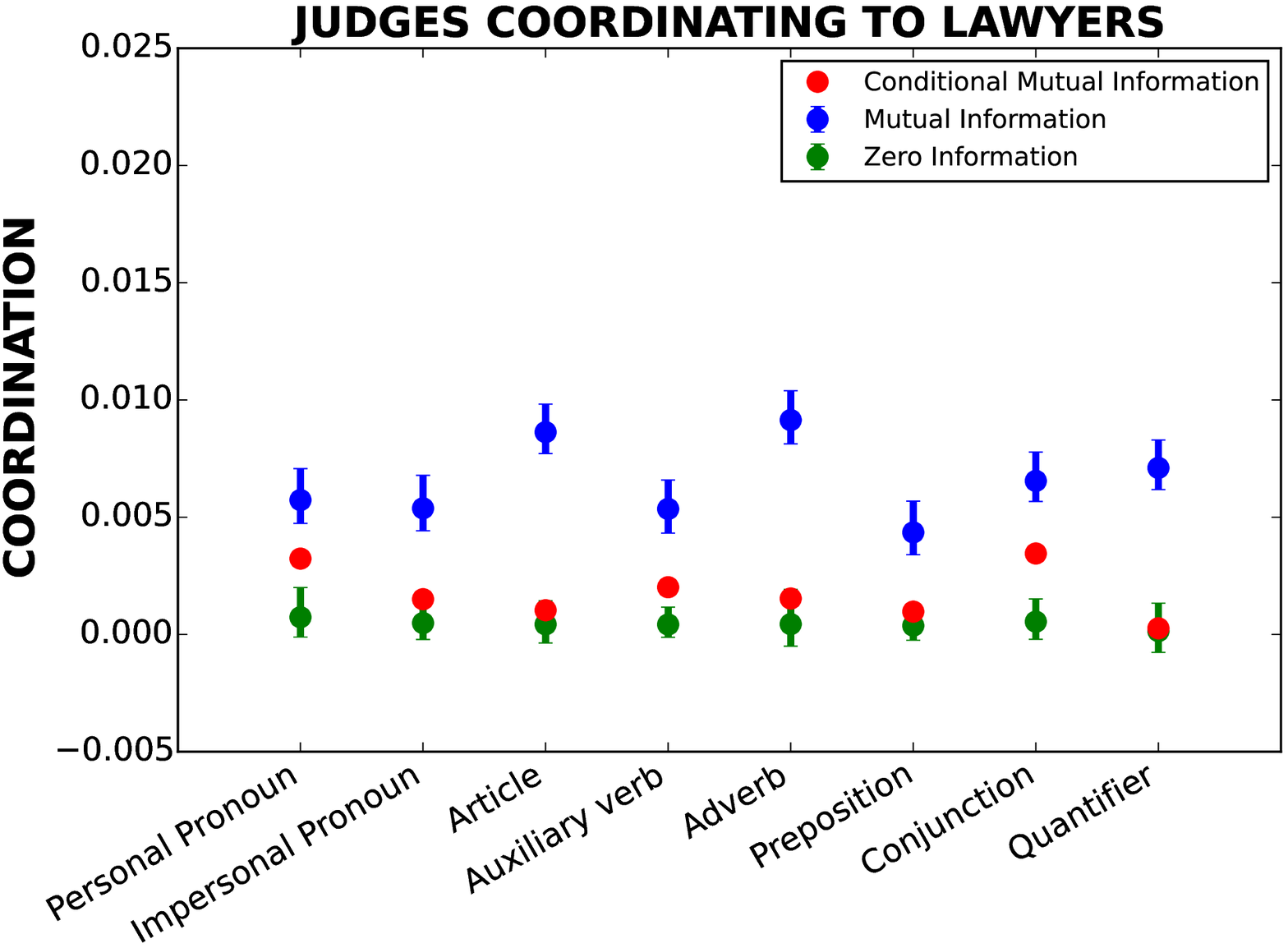}
 \label{fig_supreme_court_style_b}
  }
     \caption{ 
     \revise{\textbf{Coordination measures for the Supreme Court data.} The red (blue) dots give the true CMI (MI). The green dots represent CMI under the null hypothesis that there is no coordination after conditioning. (a) Lawyers coordinating to Judges. (b) Judges coordinating to Lawyers. In both figures, the conditional mutual information is significantly smaller than the mutual information for all eight stylistic features, indicating length is a confounding factor.}}
     \label{fig_supreme_court_style}
    \end{figure}

    \begin{figure}[htbp]
    \centering
\subfigure[]{
    \includegraphics[width=.45\columnwidth]{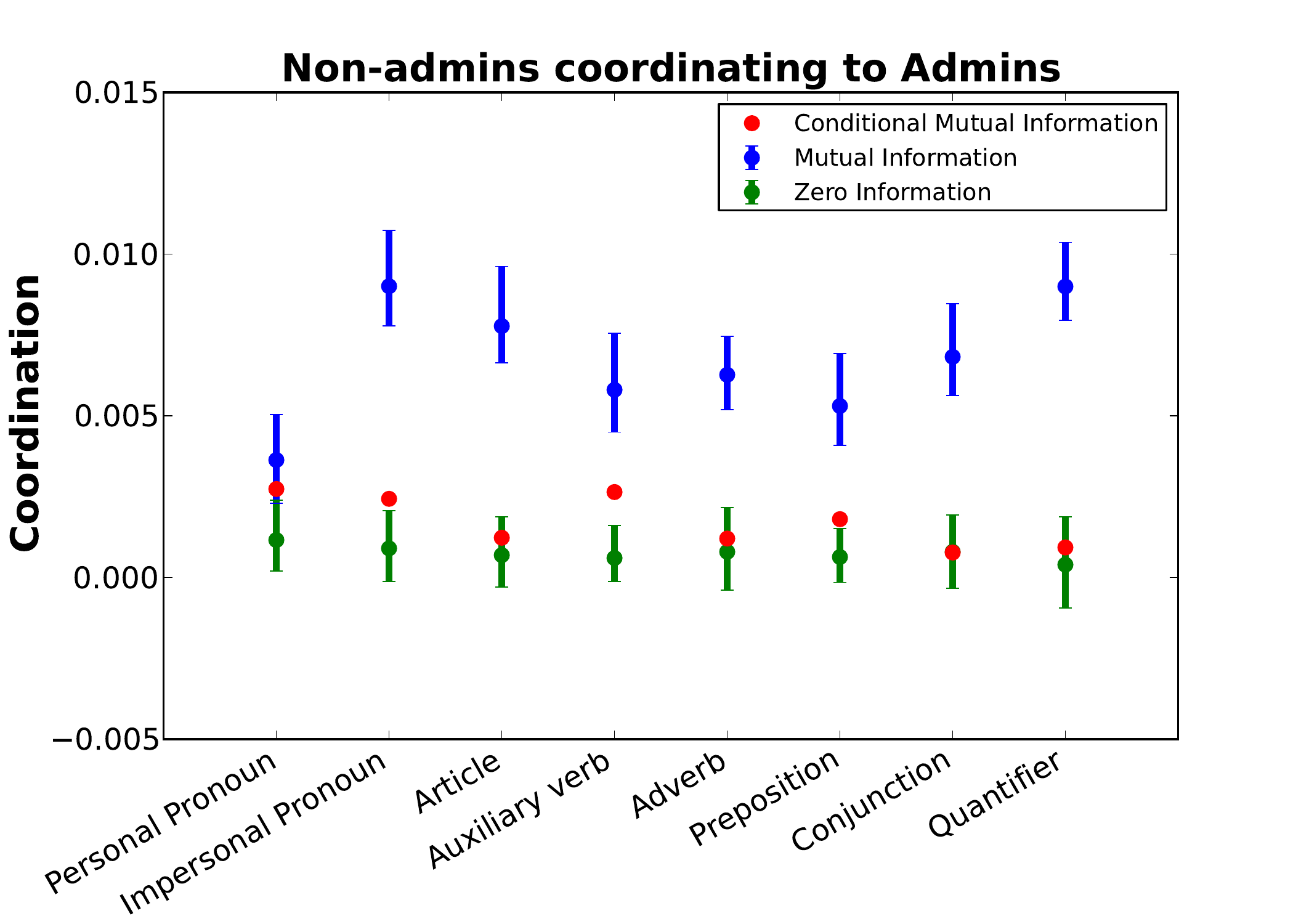} \label{fig_wiki_style_a}
    } 
    \subfigure[]{
    \includegraphics[width=.45\columnwidth]{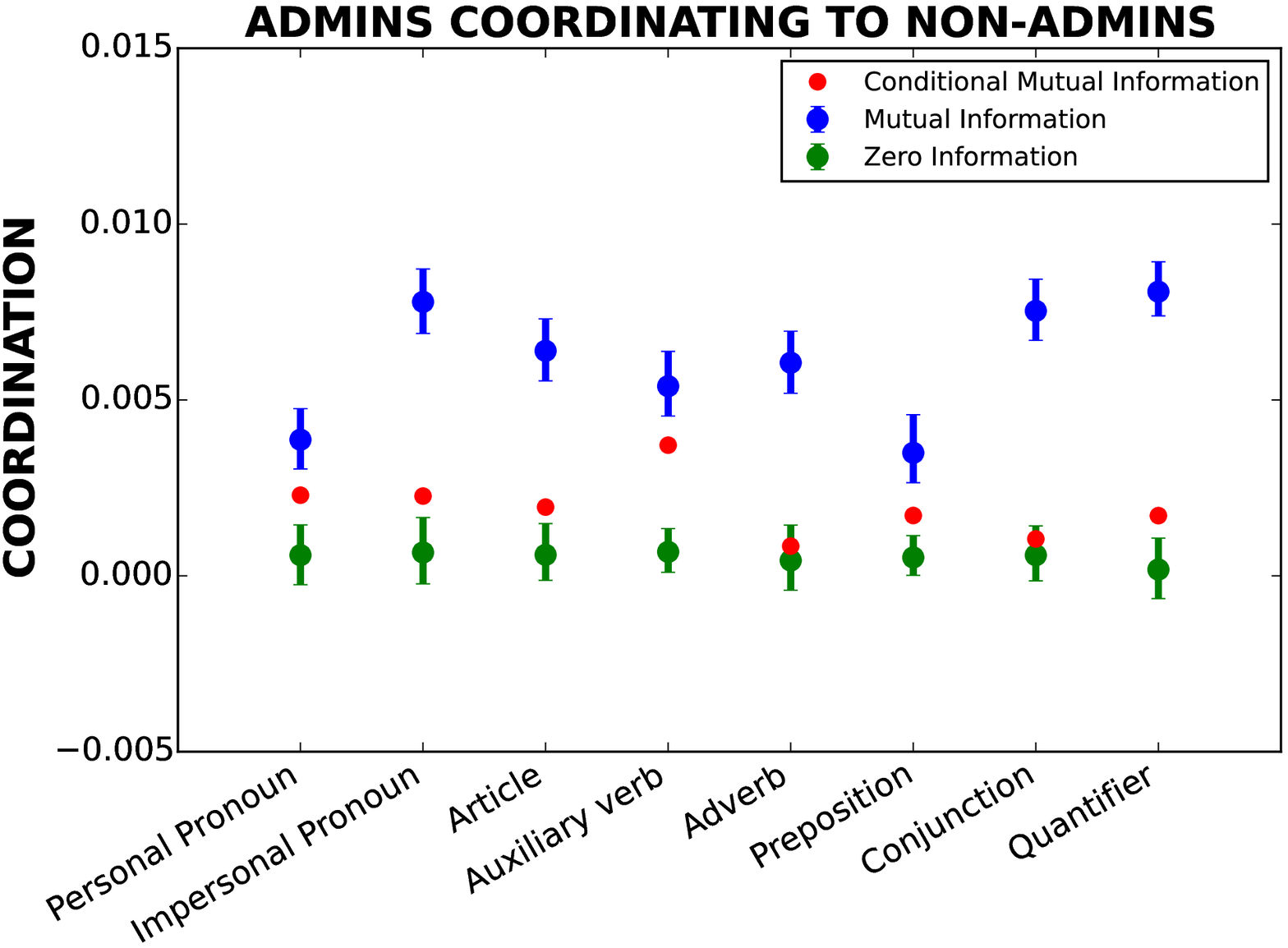} \label{fig_wiki_style_b}
    }
    
    \caption{\textbf{Coordination measures for the Wikipedia data.} (a) Non-admins coordinating to Admins. (b) Admins coordinating to Non-admins. Symbols have the same interpretation as in the previous plot.}
    \label{fig_wiki_style}
    
    \end{figure}

%
%

Fig.~\ref{fig_supreme_court_style}  describes stylistic coordination for the Supreme Court data as measured by $I(O:R)$ and $I(O:R|L_R)$. The bias in estimators for conditional mutual information and mutual information are generally different. Therefore, rather than estimating mutual information directly, we use a conditional mutual information estimator where we condition on randomly permuted values for $L_R$. We repeat this procedure for \revisetwo{four hundred times} to produce $99\%$ confidence intervals for $I(O:R)$ (blue bars). The green bars give the 99\% confidence intervals in case there is no stylistic coordination by estimating CMI with \textit{R\rq{}s} utterances permuted (erasing any stylistic coordination). 

The blue dots show the mutual information between the corresponding stylistic features, and suggest strong linguistic correlations between the groups. This effect, however, is strongly diminished after conditioning on the length of utterances (red dots). For instance, the coordination scores on features {\em Impersonal Pronoun}, {\em Article}, and {\em Auxiliary Verb} are reduced by factors of  $\sim 6.7$, $\sim 4.8$, and $5.3$, respectively, after conditioning on length. \revise{For the feature {\em Conjunction}, the 99\% confidence interval of coordination score is above the confidence interval of \textit{zero information} before conditioning, and falls into the confidence interval of \textit{zero information} after conditioning.} Similarly, in Fig.~\ref{fig_supreme_court_style}(b), the coordination scores for five out of eight markers ({\em Impersonal Pronoun}, {\em Article}, {\em Adverb}, {\em Preposition}, {\em Quantifier}) become practically zero after conditioning, suggesting that  the observed coordination in those stylistic features are due to length correlations. 



A similar picture holds for the Wikipedia dataset shown in Fig.~\ref{fig_wiki_style}. Again, we observe non-zero mutual information in all the features. However, this correlation is significantly diminished after conditioning on length.  In fact, both non-admins coordinating to admins (Fig.~\ref{fig_wiki_style}(a)) and admins coordinating to non-admins (Fig.~\ref{fig_wiki_style}(b)) have an extremely weak  signal after conditioning on length (all below 0.005). In particular, for non-admins coordinating to admins(Fig.~\ref{fig_wiki_style}(a)), the red dots of five out of eight features lie in the zero conditional mutual information confidence interval. For these five features in Fig.~\ref{fig_wiki_style}(b), we cannot rule out the null hypothesis that all stylistic coordination is due to phenomenon of length coordination.  

Another interesting observation is that there is significant asymmetry, or {\em directionality}, in stylistic coordination. For instance, by comparing Figs.~\ref{fig_supreme_court_style}(a) and~\ref{fig_supreme_court_style}(b) we see that the mutual information is significantly higher from lawyers to judges than vice versa. A similar (albeit less pronounced) asymmetry is present for the Wikipedia data as well. This type of asymmetry has been used to suggest that the relative strength of stylistic accommodation reflects social status~\cite{Danescu-Niculescu-Mizil2012WWW}. However, Figs.~\ref{fig_supreme_court_style} and~\ref{fig_wiki_style} illustrate that the asymmetry is drastically  weakened after conditioning on length (red dots), suggesting that the phenomenon of higher stylistic coordination  from lawyers to judges (and from non-admins to admins for the Wikipedia dataset) is due to the confounding effect of length. Unfortunately, a direct assessment of this effect in a single conversation is not feasible due to the insufficient number of utterances for calculating conditional mutual information. Nevertheless, in \revisethree{S2 File} we suggest a different approach for addressing the above problem,  and find that asymmetry in stylistic coordination can be explained by asymmetry in length coordination.

To conclude this section, we note  that some of the correlations in stylistic features persist even after conditioning on length. One can ask whether this remnant correlation is due to turn-by-turn level linguistic coordination, or can be  attributed to other confounding factors. We address this question in detail later in the text.

\section*{Understanding Length Coordination}\label{ULC}

As discussed in the previous section, the observed correlations in linguistic features can be attributed to coordination in the length of utterances. Here we analyze this phenomenon in more detail. In particular, we are interested whether the observed length correlations are due to turn-by-turn coordination, or can be attributed to other contextual factors. For instance, consider a scenario that in one conversation, {\em Alice} and {\em Bob} are always conversing using short statements, while in another conversation they exclusively use long statements, perhaps due to different topics of conversation. Length coordination is found if data from these two conversation is aggregated, however, this coordination only reflects {\em Alice}'s and {\em Bob}'s response to the topic of conversation. 
More generally, aggregating data might lead to effects similar to Simpson's paradox~\cite{Simpson1951}.

\begin{figure}[htbp]
\centering
\includegraphics[width=0.3\columnwidth]{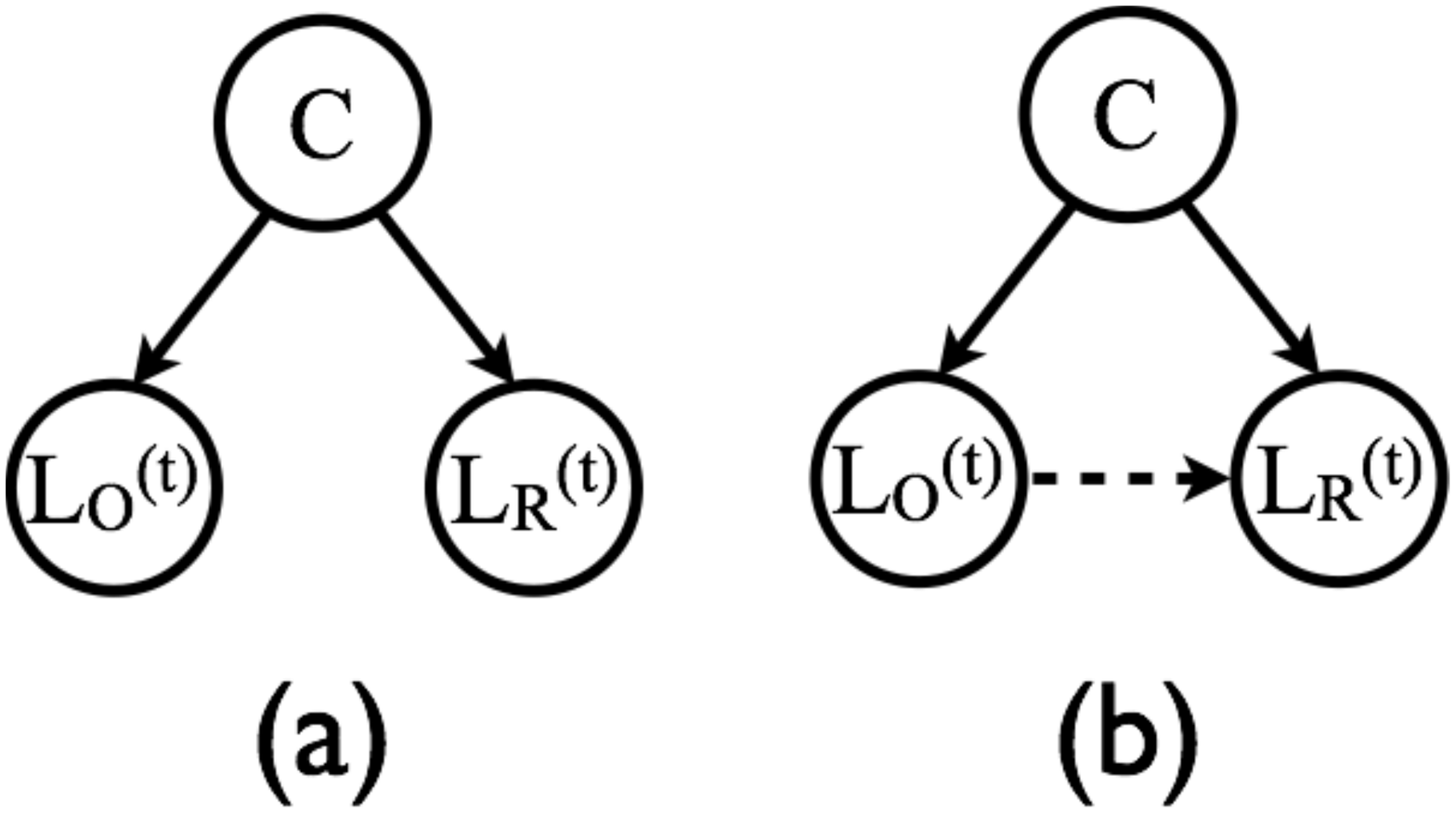} 
\caption{ \textbf{A Bayesian network model for length coordination.} The network containing contextual factors, $C$, the length of an utterance, $L_O^{(t)}$, and the length of the response, $L_R^{(t)}$. \revisetwo{(a) The lengths are correlated only due to contextual factors. (b) The lengths are correlated due to both contextual factors and potential effect of turn-by-turn level coordination (represented with the dotted line).}
 }
 \label{fig_length_model}
\end{figure}

To understand the possible extent of various confounding factors (we call them \textit{contextual factors}),  consider the Bayesian network model  that incorporates both \textit{contextual factors} and \textit{length coordination}, as shown in Fig.~\ref{fig_length_model}.  Here  $L_O$ and $L_R$ are  random variables representing the length of an utterance by the originator $O$ and the respondent $R$, respectively. \revisetwo{In the model with both solid and dashed lines in Fig.~\ref{fig_length_model}(b), $L_O$ explicitly influences $L_R$. While if we only have the soild lines in Fig.~\ref{fig_length_model}(a), $L_R$ is independent of $L_O$ after conditioning on the context $C$. Thus, the model in Fig.~\ref{fig_length_model}(a) assumes that there is only contextual coordination while Fig.~\ref{fig_length_model}(b) implies turn-by-turn coordination.} \revisethree{Note that in principle, the contextual factor $C$ can vary within a single conversation, for example, the theme of a conversation may change as time goes by. But for simplicity, we will assume that the contextual factor $C$ does not change within the dialogue or conversation.}

%
%
%
\subsection*{Information-theoretic characterization of length coordination} 

A direct measure of Turn-by-turn Length Coordination ($TLC$) is given by the following conditional mutual information:
\BEQ
TLC = I(L_O:L_R|C)
\label{eq_TLC}
\EEQ
Additionally,  we define the Overall Length Coordination ($OLC$) as
 \BEQ
 OLC = I(L_O:L_R) 
 \EEQ
Thus, $OLC$  captures not only the length coordination in a turn-by-turn level, but also the confounding behaviors between $L_O$, $L_R$ and $C$. In fact, $OLC$ can be decomposed into two items:
 \BEQ
 OLC = TLC + I(L_O:L_R:C)
 \label{eq_OLC} 
 \EEQ
 The second item of right hand side in Eq.~\ref{eq_OLC} indicates the multivariate mutual information(MMI) (also known as \textit{interaction information}~\cite{McGill54} or \textit{co-information}~\cite{bell03}), and characterizes the amount of shared information between $L_O$, $L_R$ and $C$. 

A straightforward method to test for turn-by-turn coordination is to evaluate  $TLC$ described in Eq.~\ref{eq_TLC}. Indeed, $L_O$ and $L_R$ are conditionally independent of $C$ if and only if $TLC=0$. However, direct evaluation of $TLC$ is not possible due to the lack of sufficient number of samples, e.g., the number of exchanges within a specific dialogue. Nevertheless, it is possible to test the turn-by-turn length coordination by a non-parametric statistical test as shown below. 

\subsection*{Turn-by-Turn Length Coordination Test}  
\label{TTLCT}
\revisetwo{
Our null hypothesis is that there is no turn-by-turn coordination, so that all observed correlations are due to contextual factors. We now describe a procedure for testing this hypothesis. 

We denote the pairwise set of exchanges in a specific dialogue $c$ from originator $o$ and respondent $r$ as:
\BEQ
{S_{o \leftarrow r}^c} = \left\{ {o_c^k,r_c^k} \right\}_{k = 1}^{{K_c}}
\label{pairwise_ex}
\EEQ
where ${o_c^k}, {r_c^k}$ indicate the $k$th exchange (two utterances) by the originator $o$ and respondent $r$ in dialogue $c$, and $K_c$ represents the total number of exchanges in $c$. We also define the aggregated set of exchanges of user $o \in O$ and user $r \in R$ as:
\BEQ
{S_{O \leftarrow R}} = \bigcup\limits_{o \in O,r \in R} {\bigcup\limits_{c \in {C_{o,r}}} {{S_{o \leftarrow r}^c}} } 
\EEQ
where $C_{o,r}$ represents all the dialogues that involved user $o$ and $r$. 
We can rewrite $S_{O \leftarrow R}$ element-wise as 
\BEQ
{S_{O \leftarrow R}} = \left\{ {{O_k},{R_k},{C_k}} \right\}_{k = 1}^N
\label{eq_S}
\EEQ
where $N=\left| {{S_{O \leftarrow R}}} \right|$ representing number of samples. For each triplet of right hand side in Eq.~\ref{eq_S}, $R_k$ is the reply utterance to $O_k$ in the dialogue $C_k$. Finally, from $S_{O \leftarrow R}$ we obtain the set 
\BEQ
L( {{S_{O \leftarrow R}}} ) = \left\{ {len\left( {{O_k}} \right),len\left( {{R_k}} \right)} \right\}_{k = 1}^N
\label{length_sample}
\EEQ 
where $len\left(  \cdot   \right)$ is a function representing the length of an utterance.

Consider now another sample, which is obtained by randomly permuting the respondent $r$'s utterances in the set $S_{o \leftarrow r,c}$: 
\BEQ
\widehat S_{o \leftarrow r}^c = \left\{ {o_c^k,\widehat r_{c}^{k}} \right\}_{k = 1}^{{K_c}}
\EEQ 
where $\{ {\widehat r}_c^k  \}_{k = 1}^{{K_c}}$ is a random permutation of $\left\{ {r_c^k} \right\}_{k = 1}^{{K_c}}$. By aggregation, we have,
\BEQ
\widehat S_{{{O \leftarrow R}}} = \bigcup\limits_{o \in A,r \in B} {\bigcup\limits_{c \in {C_{o,r}}} {\widehat S^c_{{{o \leftarrow r}}}} }  = \left\{ {{O_k},\widehat R_k,{C_k}} \right\}_{k = 1}^N \nonumber
\label{permute_set}
\EEQ
and
\BEQ
L( {\widehat S_{O \leftarrow R}} ) = \{ {len ( {{O_k}} ),len( {\widehat R_{{k}}} )} \}_{k = 1}^N
\label{length_permute}
\EEQ
Let us assume there is no turn-by-turn coordination, so that $L_O$ and $L_R$ are conditionally independent from each other given $C$. Then, it is easy to see that under this null model, the samples $L( {{S_{O \leftarrow R}}} )$ and $L( {\widehat S_{O \leftarrow R}} )$ have the same likelihood, e.g., they are statistically equivalent. In other words, $L( {\widehat S_{O \leftarrow R}} )$  can be viewed as a new sample from the same distribution $p( l_o,l_r)$. This observation suggests the following test: We first estimate $OLC$ from the sample $L( { S_{O \leftarrow R}} )$ (denoted as $OLC_0$) and then using the within-dialogue shuffled samples $L( {\widehat S_{O \leftarrow R}} )$ (denoted as $OLC_1$). Under the null hypothesis, these two estimates should coincide. Conversely, if $OLC_0\neq OLC_1$, then the null hypothesis is rejected, suggesting that there is turn-by-turn length coordination. 
}

 The above procedure, which we call Turn-by-Turn Length Coordination Test,  is a conditional Monte Carlo test~\cite{Pesarin_Book}. The main advantage of this non-parametric test is that it requires a smaller sample size and does not need to make particular distribution assumptions. The test is non-parametric in two ways: the permutation procedure is non-parametric as well as the estimation of mutual information. We also note that in the context of stylistic coordination, a similar test was used in Ref.~\cite{Danescu-Niculescu-Mizil2011CIC}.

%

\begin{figure}[htbp]
\centering
\subfigure[]{
    \includegraphics[width=0.45\columnwidth]{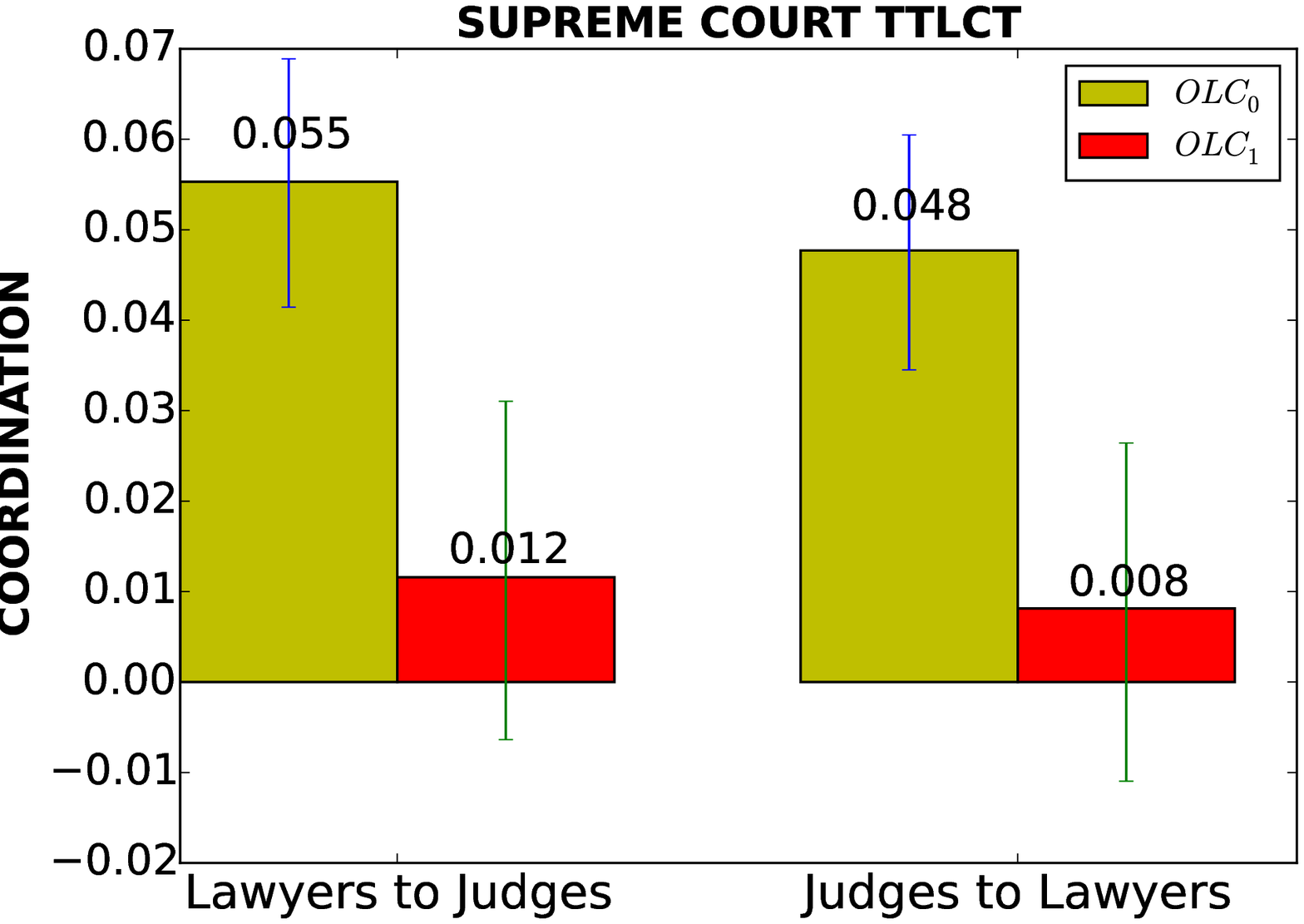} \label{fig_length_supreme_court}
    } 
    \subfigure[]{
    \includegraphics[width=0.45\columnwidth]{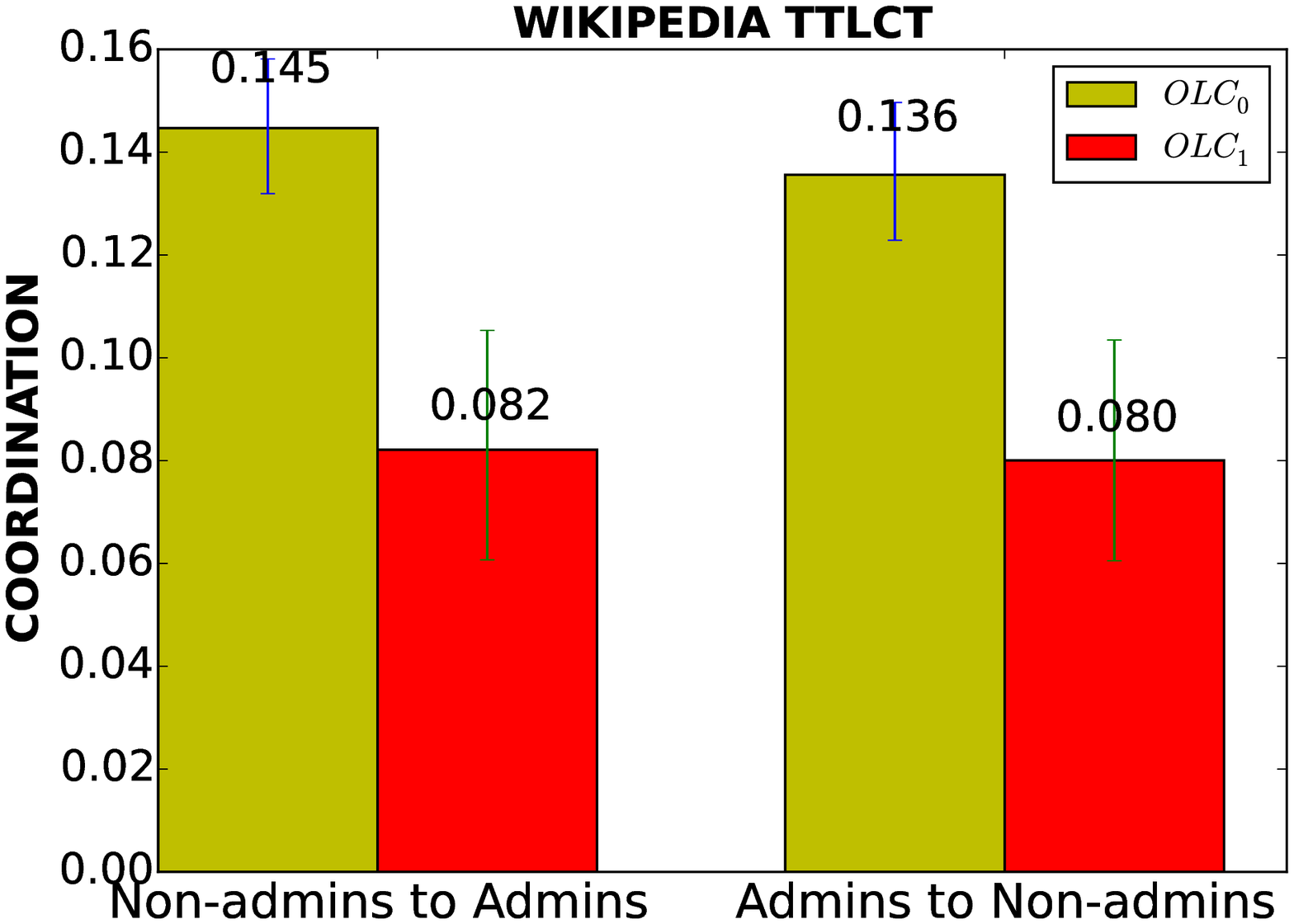} \label{fig_length_wiki}
    }

\caption{\revise{\textbf{Turn-by-turn length coordination test.} (a) Supreme Court dataset. (b) Wikipedia dataset. In both two subfigures, $OLC_1$ is significantly smaller than $OLC_0$.}  
\label{fig:Lcoord}
}
\end{figure}
The results of this test are shown in Fig.~\ref{fig:Lcoord}. For the Supreme Court data, Fig.~\ref{fig:Lcoord}(a) shows that both Lawyers to Judges and Judges to Lawyers have non-zero mutual information ($OLC_0$) before permutation. The  Turn-by-Turn Length Coordination test shows that the mutual information decreases significantly after permutation(green confidence intervals, $OLC_1$), rejecting the null hypothesis that $L_O$ and $L_R$ are independent after conditioning on the contextual factor $C$. In other words, the contagion of length exists from the original utterance to the reply on a turn-by-turn level.

%

 For the results on the Wikipedia discussion board in Fig.~\ref{fig:Lcoord}(b),  we are also able to reject the null hypothesis. Notice that the degree of mutual information $OLC$ is higher for Wikipedia than for the Supreme Court. However, one cannot make a general conclusion about the exact magnitude of turn-by-turn length coordination ($TLC$) simply by calculating the loss, i.e., $OLC_0-OLC_1$.

\section*{Revisiting Stylistic Coordination}
\label{sec:revisit}
 
 We demonstrated in the previous section that strong correlations in utterance length explain most of the observed stylistic coordination. However, in some situations, there are statistically significant non-zero signals even after conditioning on length, e.g., the first feature dimension {\em Personal Pronoun} in Figs.~\ref{fig_supreme_court_style}(a) and~\ref{fig_supreme_court_style}(b). 
 We now proceed to  examine this remnant coordination. Specifically, we are interested in the following question: Does the non-zero conditional mutual information (after conditioning on length) represent  turn-by-turn level stylistic coordination, or is it due to other contextual factors?

\begin{figure}[htbp]

\centering
\includegraphics[width=0.3\columnwidth]{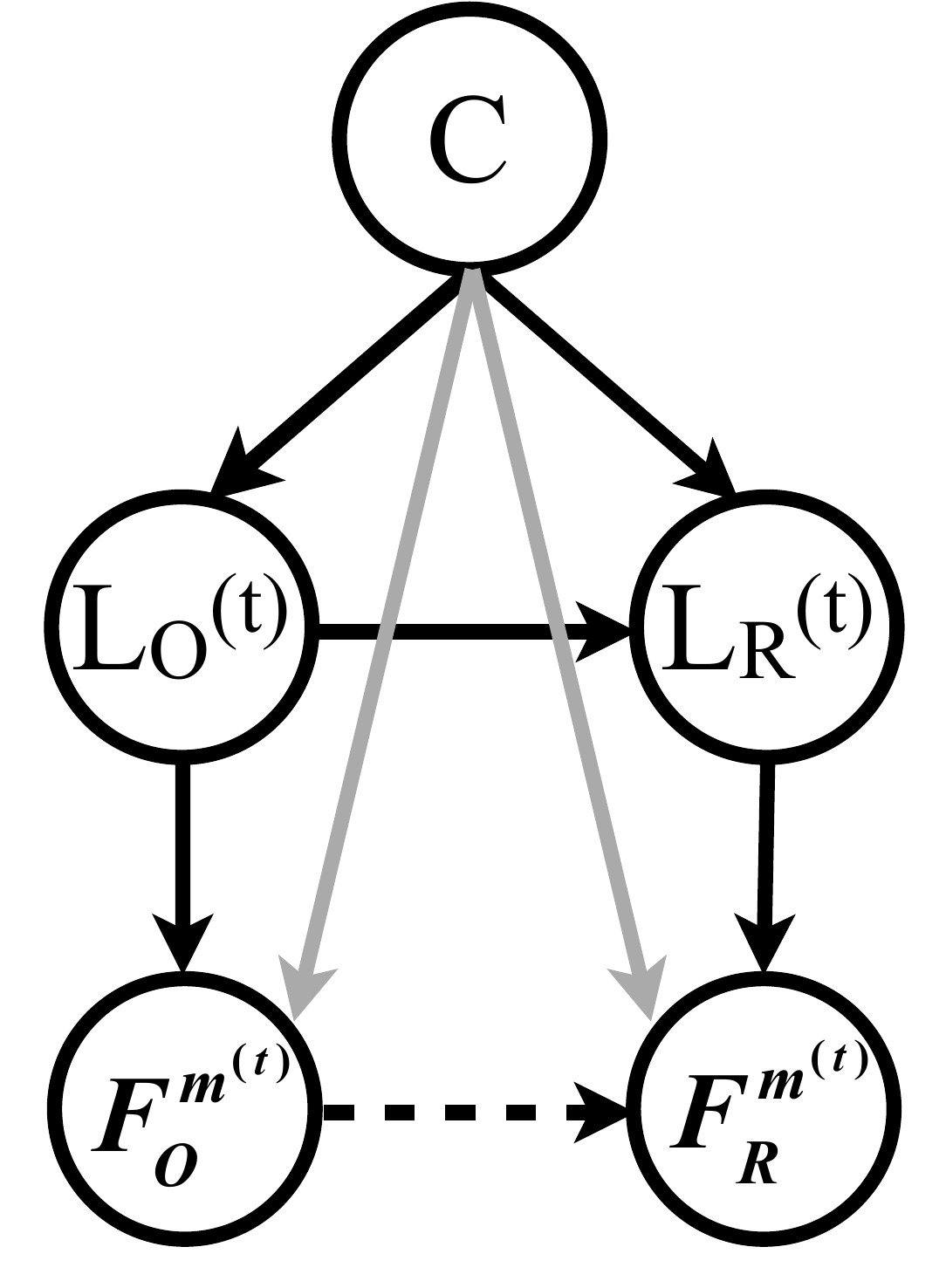}
\caption{ \textbf{A Bayesian network for linguistic style coordination.} $L_O$ and $L_R$ represent length of the respondent and length of the originator respectively. $F^m_O$ and $F^m_R$ represent a specific style feature variable for the respondent and originator. 
     }
\label{fig_length_style_model}
\end{figure}

Toward this goal, consider the  Bayesian network in Fig.~\ref{fig_length_style_model}, which depicts conditional independence relations between the length variables $L_O$ and $L_R$; stylistic variables $F^m_O$ and $F^m_R$ with respect to a style feature $m$, and the contextual (dialogue) variable $C$. The solid arrow from $L_O$ to $L_R$ reflects our findings from the last section about the existence of turn-by-turn length coordination. The dashed arc between the features $F_O^m$ and $F_R^m$ characterizes turn-by-turn stylistic coordination. Finally, the grey  arcs between $C$, $F_O^m$ and $C$, $F_R^m$ indicate possible contextual coordination. 

We use conditional mutual information to measure the Turn-by-turn Stylistic Coordination (TSC) with respect to a specific style feature $m$:
\BEQ
TSC = I(F^m_O:F^m_R|C, L_R)
\label{eq_TSC}
\EEQ
where $F^m_O$, $F^m_R$ are binary variables indicating the feature $m$ appears or not in an utterance. Also, the Overall Stylistic Coordination(OSC) is defined as
 \BEQ
 OSC = I(F^m_O;F^m_R|L_R) 
 \label{eq_OSC}
 \EEQ
 Thus, $OSC$ is exactly the conditional mutual information introduced in Eq.~\ref{eq:cmi}. Note that, even after conditioning on length, $F^m_O$ and $F^m_R$ are still dependent of each other because they are sharing the contextual factor $C$. ($F^m_O \leftarrow C \rightarrow F^m_R$ is called a \textit{d-connected} path in~\cite{Pearl2000})

 Again, a direct measure of turn-by-turn stylistic coordination corresponds to non-zero $TSC$ in Eq.~\ref{eq_TSC}. However, $TSC$ is hard to evaluate due to lack of sufficient samples. Furthermore, the  shuffling test from the previous sections is not directly applicable here either, because it needs to be done in way that keeps the correlations between $L_O$ and $L_R$ intact: In other words, one can exchange utterances that have the same lengths. Since most dialogues are rather short, this type of shuffling test is not feasible, and one needs an alternative approach. 
 

 
 \subsection*{Turn-by-Turn Stylistic Coordination Test} \label{TTSCT}

 Our proposed test is based on the following idea: if we can rule out the influence of the contextual factors on stylistic correlations, then any non-zero conditional mutual information can be only explained by turn-by-turn stylistic coordination, i.e., $OSC=TSC$. Thus, the null hypothesis is that there is contextual level coordination in stylistic features. We emphasize that by contextual coordination, we are actually referring to the links from $C$ to $F_O$ and $C$ to $F_R$ in Fig.~\ref{fig_length_style_model}. 

We follow the same notation and methodology used in previous sections. By Eq.~\ref{eq_S}, let us denote the mixed length and stylistic feature set of ${S_{O \leftarrow R}}$ as:
\revisetwo{
\BEQ
L{F_m}\left( {{S_{O \leftarrow R}}} \right) = \left\{len\left( {{O_k}} \right),len\left( {{R_k}} \right), {{f_m}\left( {{O_k}} \right),{f_m}\left( {{R_k}} \right),{C_k}} \right\}  \nonumber 
\label{eq_LF} 
\EEQ
}
where ${f_m}\left(  \cdot  \right)$ is a binary function represents whether the style feature $m$ in an utterance appears or not.



Consider now the shuffling procedure: we randomly permute respondent's utterances within a dialogue and obtain the set $\widehat S_ {{{O \leftarrow R}}}$ in Eq.~\ref{permute_set}.  We also define the length and feature set of $\widehat S_{O \leftarrow R}$: 
\revisetwo{
\BEQ
L{F_m}( {\widehat S_{{{O \leftarrow R}}}} ) = \{ {len ( {{O_k}} ),len ( \widehat R_k  ),{f_m} ( O_k),  f_m ( \widehat R_k ),{C_k}} \} \nonumber 
\label{eq_LF_shuffled}
\EEQ
}
Clearly,  the permutation destroys the turn-by-turn level coordination in both length and style. Thus, any remnant correlation must be due to contextual coordination, e.g., the fork $F^m_O \leftarrow C \rightarrow F^m_R$. This provides a straightforward test for the existence of contextual coordination. Indeed, we simply need to estimate the overall stylistic coordination $OSC_1$ using the shuffled sample $L{F_m}( {\widehat S_{{{O \leftarrow R}}}} )$. If $OSC_1$ is larger than zero, then there is necessarily contextual coordination. On the other hand, if $OSC_1=0$, then all the observed stylistic correlations (calculated using the original non-shuffled sample) must be due to turn-by-turn stylistic coordination.

\begin{figure}[htbp]
\centering
\subfigure[]{
    \includegraphics[width=0.45\columnwidth]{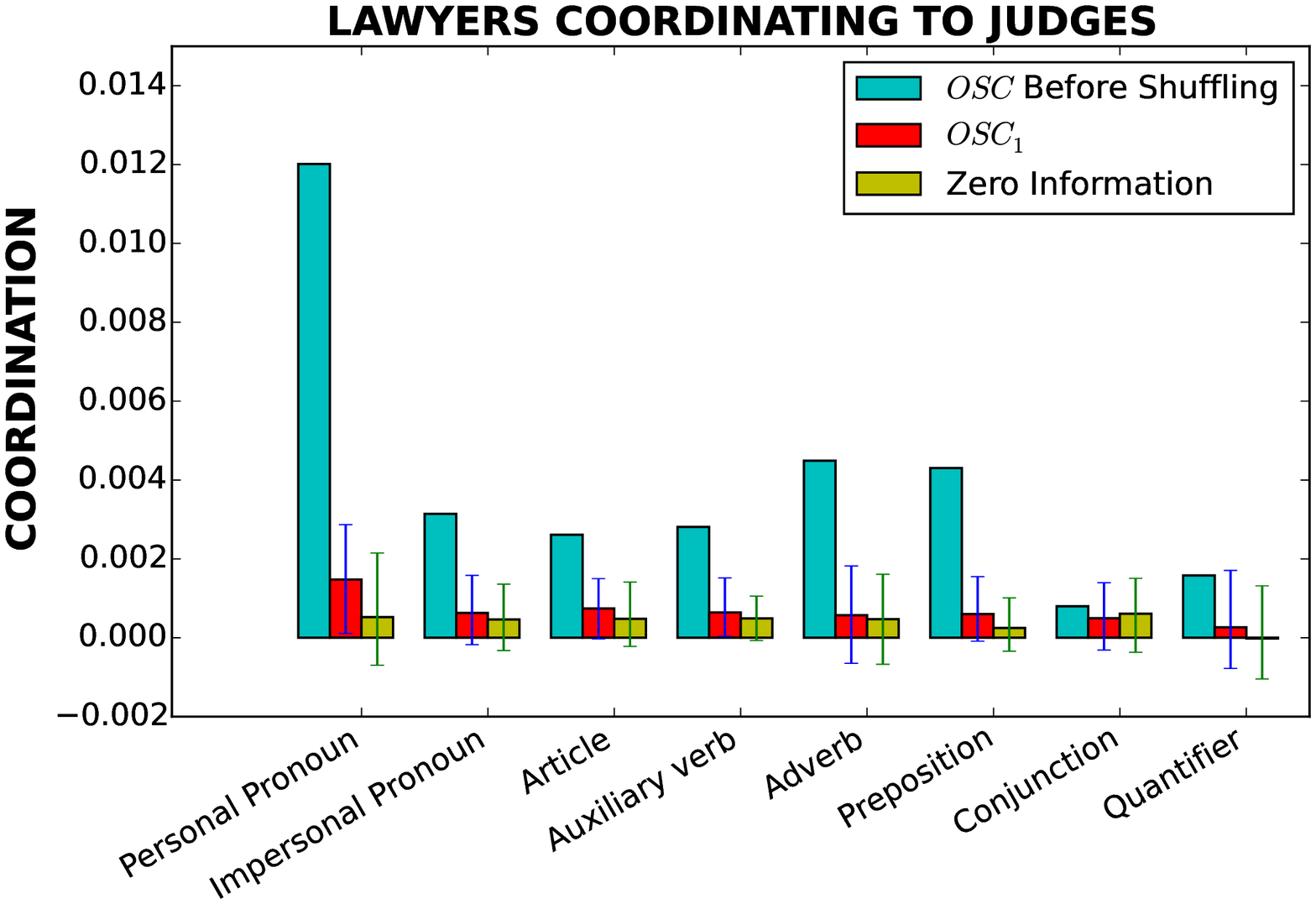} \label{fig_supreme_court_turn_by_turn_a}
 
    } 
    \subfigure[]{
    \includegraphics[width=0.45\columnwidth]{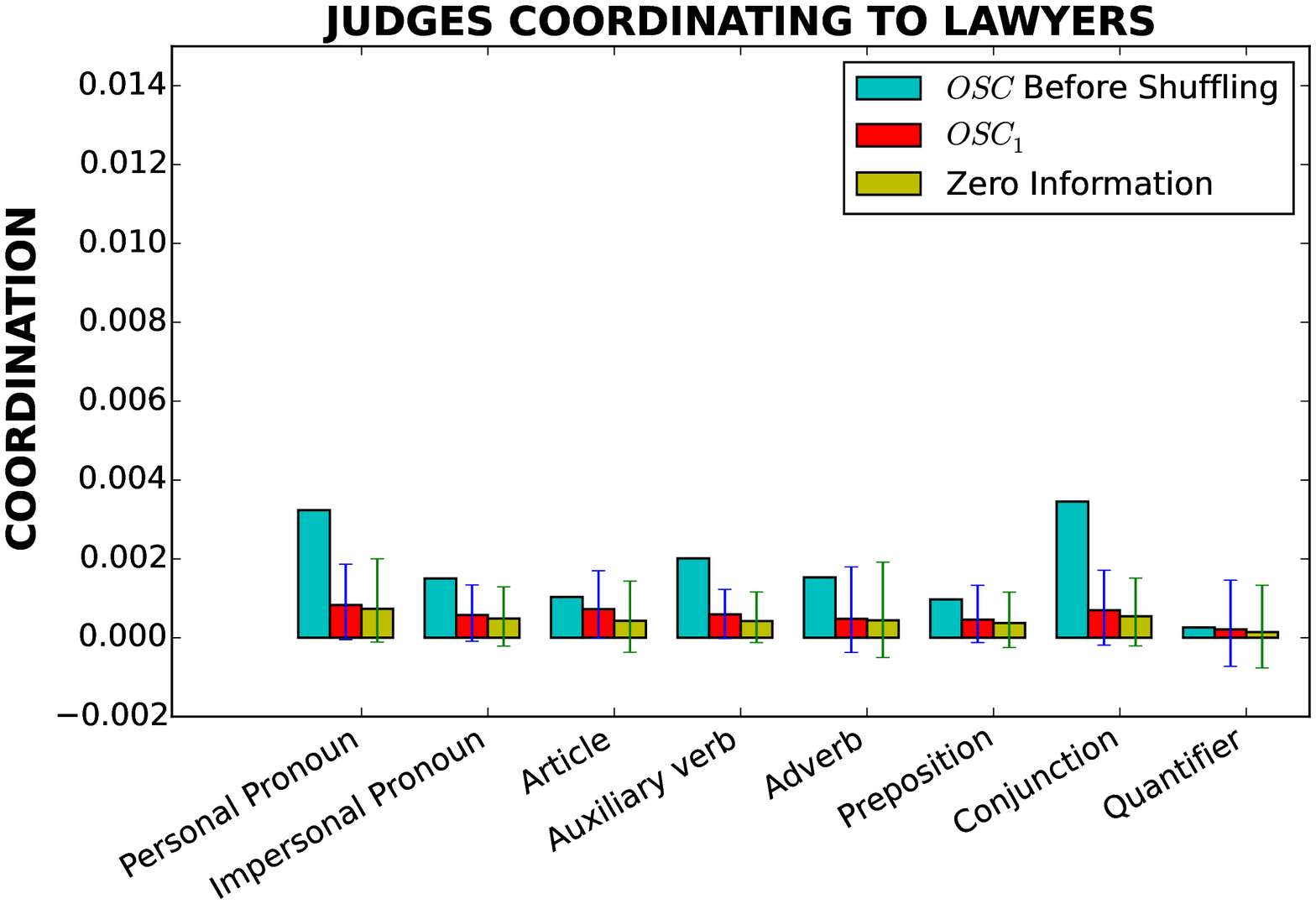} \label{fig_supreme_court_turn_by_turn_b}
    }
    
    \caption{\revise{\textbf{Turn-by-turn stylistic coordination test for Supreme Court data.} (a) Lawyers coordinating to Judges. (b) Judges coordinating to Lawyers.  (Blue bars indicate the overall stylistic coordination($OSC$) before the test). One can see that after shuffling, values of $OSC_1$ are within the zero-information confidence intervals.}}
    \label{fig_supreme_court_turn_by_turn}
\end{figure}
Let us first consider the results of the above test for the Supreme Court data. From Figs.~\ref{fig_supreme_court_turn_by_turn}(a) and~\ref{fig_supreme_court_turn_by_turn}(b), one can see that for all the features, the corresponding CMI $OSC_1$ are within the zero-information confidence intervals, indicating that non-zero conditional mutual information  ($OSC$ before shuffling) cannot be attributed to contextual factors. In other words, the remnant correlations that are not explained by length coordination must be due to turn-by-turn level coordination. 

\begin{figure}[htbp]
\centering
\subfigure[]{
    \includegraphics[width=0.45\columnwidth]{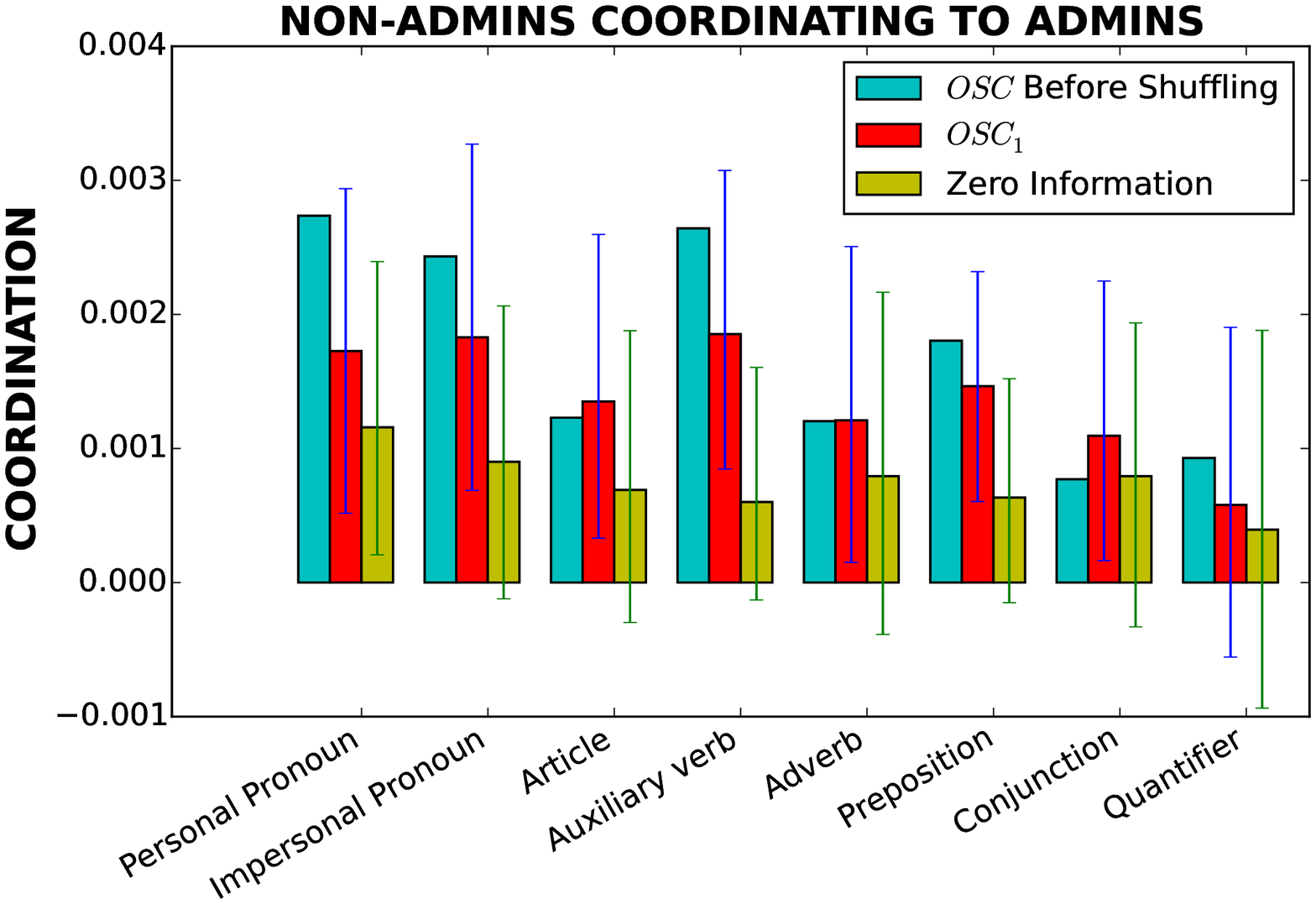} \label{fig_wiki_turn_by_turn_a}
 
    } 
    \subfigure[]{
    \includegraphics[width=0.45\columnwidth]{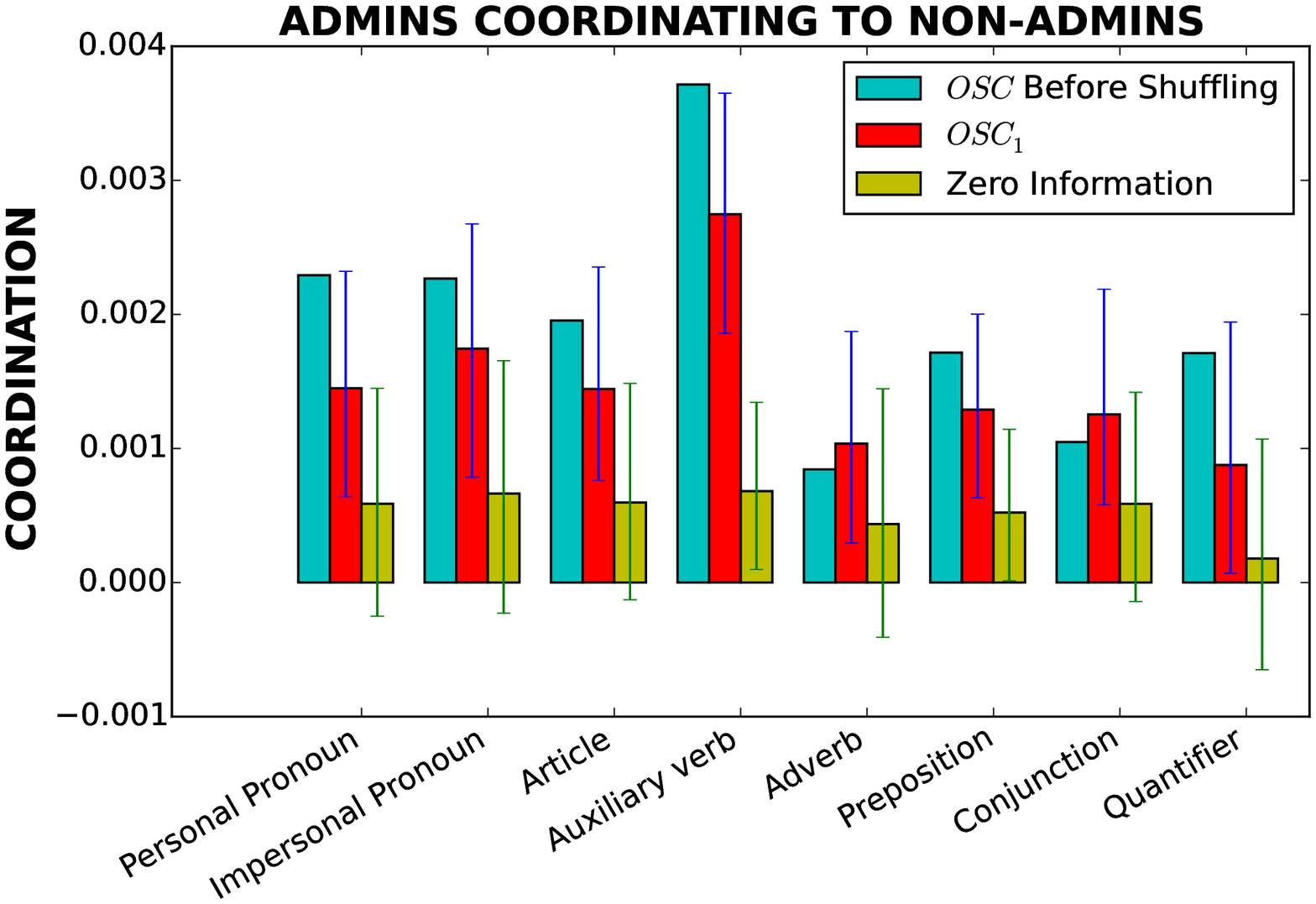} \label{fig_wiki_turn_by_turn_b}
   	 }
	\caption{\revise{\textbf{Turn-by-turn stylistic coordination test for Wikipedia data.} (a) Non-admins coordinating to Admins. (b) Admins coordinating to Non-admins. (Blue bars indicate the overall stylistic coordination($OSC$) before the test). One cannot rule out the null hypothesis that the remnant stylistic coordination is due to the contextual factors.}}
        \label{fig_wiki_turn_by_turn}
\end{figure}
\revise{The situation is different for Wikipedia data. Indeed, Figs.~\ref{fig_wiki_turn_by_turn}(a) and~\ref{fig_wiki_turn_by_turn}(b) show that for the stylistic features with statistically significant remnant correlations even after conditioning on length ($OSC$), the results of the above permutation tests are rather inconclusive. Namely,  although the confidence intervals of $OSC_1$ do overlap with the zero information confidence intervals, one cannot state unequivocally that they are zero. In other words, one cannot rule out the null hypothesis that the remnant stylistic coordination is due to the contextual factors rather than turn-by-turn coordination.} 

\section*{Discussion}

In conclusion, we have suggested an information theoretic framework for measuring and analyzing stylistic coordination in dialogues. \revisetwo{We first extracted the stylistic features from the dialogue of two participants and then used \textit{Mutual Information(MI)} as a theoretically motivated measure of dependence to characterize the amount of stylistic coordination between the originator and the respondent in the dialogue. Moreover, by introducing \textit{Conditional Mutual Information(CMI)}, which allows us to measure the correlation between two variables \textit{after} conditioning on a third variable, we are able to more accurately gauge stylistic accommodation by controlling for confounding effects like length coordination.}

\revisetwo{We then used the proposed method to revisit some of the previous studies that had reported strong stylistic coordination}. While the suggestion that one person's use of, e.g., prepositions will (perhaps unconsciously) lead the other to use more prepositions is fascinating, our results indicate that previous studies have vastly overstated the extent of stylistic coordination. In particular, we showed that a significant part of the observed stylistic coordination can be attributed to the confounding effect of length coordination. \revise{We find that for both Supreme Court and Wikipedia data, the coordination score is greatly diminished after conditioning on length. We also find that the significant asymmetry in stylistic coordination shown in the previous study~\cite{Danescu-Niculescu-Mizil2012WWW} is drastically weakened after conditioning on length. In fact, our results indicate that the asymmetry in length coordination can explain almost all the observed asymmetry in stylistic coordination.} 


\revisetwo{Simpson's paradox provides a famous example of how correlations observed in a population can disappear or even be reversed after conditioning on sub-populations. 
In an information-theoretic framework setting, a similar ``paradox'' can be seen in the example illustrated by Fig.~\ref{fig_length_model}: for $L_O$, $L_R$ and $C$, the mutual information $I(L_O:L_R)>0$, while $I(L_O:L_R|C)=0$. If we only look at the aggregated data, averaging over all contexts, $C$, i.e., $I(L_O:L_R)$, there will be artificial mutual information between $L_O$ and $L_R$. Ideally, we could calculate $I(L_O:L_R|C)$ directly, however, there may not be enough samples for us to calculate the conditional mutual information for all values of $C$. How can we still determine whether $I(L_O:L_R|C)$ is zero or not while using all the data? We thus designed non-parametric statistical tests to solve this problem in general while making full use of the available data. 
More importantly, because these information-theoretic quantities directly reflect constraints on graphical models, the mystery of Simpson-like paradoxes is replaced with concrete alternatives for generative stories as depicted in Figs.~\ref{fig_length_model} and~\ref{fig_length_style_model}.}

\revise{We also observed that for some of the stylistic markers, there was diminished but still statistically significant correlations even after conditioning on length. We again designed a non-parametric statistical test for analyzing this remnant coordination more thoroughly. Our findings suggest that for the Supreme Court data, the remnant coordination cannot be fully explained by other contextual factors. Instead, we postulate that the remnant correlations in the Supreme Court data is due to turn-by-turn level coordination. For the Wikipedia data, however, our results are less conclusive, and we cannot draw any conclusion about turn-by-turn stylistic coordination. Thus, caution must be taken when making general claims about the possible origin of stylistic coordination in different settings.}

It is possible to develop alternative tests based on a more fine-grained, token-level generative models. The main idea behind such a test is to shuffle the word tokens uttered by an individual within each dialogue, which should destroy turn-by-turn coordination. Our preliminary results based on this test suggest that most of the remnant correlations are indeed due to turn-by-turn coordination. However, we emphasize that this test requires an additional assumption whose validity needs to be verified, namely that the words used by a given speaker within a conversation are independent and identically distributed (i.i.d.). Furthermore, the test assumes stationarity, i.e.,  that the contextual factors do not vary within the course of the dialogue.  \revise{While this assumption seems reasonable in the dialogue settings considered here, it is important to note that deviations from stationarity might be yet another serious obstacle for identifying stylistic influences~\cite{win2013,steeg2012statistical}. Indeed, if we relax the stationarity condition, then any observed correlation in stylistic features might be due to temporal evolution rather than direct influence. And since any permutation-based test destroys temporal ordering, it cannot differentiate between those two possibilities.}


\revise{While our work focuses on linguistic style matching, we believe that the information-theoretic method proposed can be useful for studying more general types of linguistic coordination in dialogues, such as structural priming~\cite{bock1986syntactic,pickering2008structural}, or lexical entrainment~\cite{brennan96lexical,brennan1996conceptual}. Recall that according to the structural priming hypothesis, the presence of a certain linguistic structure in an utterance affects the probability of seeing the same structure later in the dialogue. This type of turn-by-turn coordination can naturally be captured by (time-shifted) mutual information between properly defined linguistic variables. Furthermore, using the permutation tests described here, it should be possible to differentiate between {\em historical} and {\em ahistorical} mechanisms of lexical entrainment~\cite{brennan1996conceptual}. Indeed, the former mechanism assumes some type of influence/coordination between the speakers that helps them to arrive at a common conceptualization.  The ahistorical mechanism, on the other hand, assumes that the speaker's choice of each term is an independent event affected by the informativeness and availability of the term, and some other factors, which, in our terminology, is analogous to  contextual coordination. }

In a broader context, we note that sociolinguistic analysis  has been used for assessing and predicting societally important outcomes such as health behaviors, suicidal intent, and emotional well-being, to name a few examples~\cite{Pennebaker1999,Stirman01072001,Campbell2003,Rude2004,Boals01092005}. Thus, it is imperative that such predictions are based on sound theoretical and methodological principles. Here we suggest that information theory provides a powerful computational framework for testing various hypotheses, and furthermore, is flexible enough  to account for various confounding variables. Recent advances in information-theoretic estimation are shifting these approaches from the theoretical realm into practical and useful techniques for data analysis. We hope that this work will contribute to the development of mathematically principled tools that enable computational social scientists to draw meaningful conclusions from socio-linguistic phenomena.


\section*{Acknowledgments}
This research was supported in part by Defense Advanced Research Projects Agency(DARPA) grant No. W911NF--12--1--0034.


\section*{Supporting Information}

\hspace{0.2in}\textbf{S1 File. Basic Concepts in Information Theory.}

\textbf{S2 File. Stylistic Coordination and Power Relationship.} 
\appendix
\newpage
\revise{

\renewcommand\thefigure{\Alph{figure}} 

\section*{S1 Basic Concepts in Information Theory}
\label{app:info}
Consider a random variable $X$ with probability distribution $p(x)\equiv p(X=x)$. In the discrete case, the \textit{Shannon} entropy is defined as: 
\BEQ \label{eq:ent_def}
H\left( X \right) =  - \sum_x {p\left( x \right)\log p\left( x \right)} 
\EEQ 
\revisethree{Note that If $X$ is a continuous variable, the sum in Eq.~\ref{eq:ent_def} is replaced by an integral.} We often talk about the entropy $H(X)$ s quantifying our uncertainty about the random variable, so that higher entropy means more uncertain (and less predictable) $X$. In particular, entropy is zero, $H(X)=0$, if and only if $X$ is perfectly predictable.   

Let us now consider two random variables, $X$ and $Y$, and let $p(x,y)$ denote their joint distribution. The (joint) entropy for $X$ and $Y$ is defined similarly as 
\BEQ
H\left( {X,Y} \right) =  - \sum_x {\sum_y {p\left( {x,y} \right)\log p\left( {x,y} \right)} } 
\label{eq:joint_entropy}
\EEQ
We can also define conditional entropy of $Y$ given $X$ (or vice versa), as follows:
\BEQ
H\left( {Y|X} \right) =  - \sum_x p(x) {\sum_y {p\left( {y|x} \right)\log p\left( {y|x} \right)} } 
\label{eq:cond_entropy}
\EEQ

If $X$ and $Y$ are independent, then their joint entropy is the sum of individual entropies, $H(X,Y)=H(X) + H(Y)$. Indeed, note that for independent $X$ and $Y$, their joint distribution factorizes as $p(x,y)=p(x)p(y)$. Plugging this factorized distribution into Eq.~\ref{eq:joint_entropy} then yields the additivity of the joint entropy.

In case $X$ and $Y$ are not independent, the degree of their dependence can be measured by {\em mutual information} (MI), defined as follows:
\BEQ
I\left( {X:Y} \right) = H\left( X \right) + H\left( Y \right) - H\left( {X,Y} \right)
\EEQ
$I\left({X:Y} \right)$ measures the correlation between $X$ and $Y$, and equals to zero if and only if $X$ and $Y$ are independent. In essence, $I(X:Y)$ measures the amount of uncertainty reduction in $X$, if we know $Y$, and vice visa. This intuition becomes more clear if we rewrite the mutual information the in the following form:
\BEA
I\left( {X:Y} \right) &=& H\left( Y \right) - H\left( Y |X \right) \\
&=& H\left( X \right) - H\left( X|Y \right) 
\EEA
The relationships between the entropies, conditional entropies, and mutual information are captured in the Venn diagram shown in Fig.~\ref{mi_explain}.
\setcounter{figure}{0} 
\begin{figure}[]
\centering
 \includegraphics[width=0.5\textwidth]{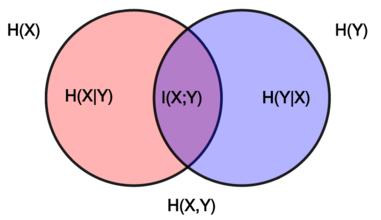}
\caption{The Venn diagram depicting the relationship between individual ($H(X)$, $H(Y)$), joint ($H(X,Y)$), and conditional ($H(X|Y)$, $H(Y|X)$) entropies. The intersection of the circles  is the mutual information $I(X;Y)$.
 }
 \label{mi_explain}
\end{figure}

For a more detailed account of information-theoretic concepts, we refer the reader to a classical textbook by Cover and Thomas~\cite{CoverThomas2006}.  
}
\newpage
  $ $
\newpage
\renewcommand\thefigure{\Alph{figure}} 
\section*{S2 Stylistic Coordination and Power Relationship}
It has been hypothesized that directionality of the stylistic coordination in dialogues can be predictive of power relationship between the conversations, as discussed in~\cite{Danescu-Niculescu-Mizil2012WWW}. We do indeed observe directional differences in stylistic coordination when comparing Figs.~\ref{fig_supreme_court_style}(a) and~\ref{fig_supreme_court_style}(b) with Figs.~\ref{fig_wiki_style}(a) and~\ref{fig_wiki_style}(b). However, as we elaborated above, the observed directionality can  result from the confounding effect of length coordination. 

Here we analyze this issue in more details by setting up the following prediction task (see~\cite{Danescu-Niculescu-Mizil2012WWW}). We consider all the pairs of users $(X,Y)$ who have different social status (e.g., admin vs. non-admin) and have engaged in dialogues. We then calculate stylistic coordination scores from $X$ to $Y$ and $Y$ to $X$, and examine whether those scores can be used to classify the social status of each speaker. For classification, we assume we know the status relationship for a fraction of pairs in our dataset, and then use a supervised learning method called Support Vector Machine (SVM) to predict the status of the unknown users. We perform this prediction tasks using the following three different set of features: 
\begin{itemize}
\item {\em Coordination Features}: For each pair, and for each of the eight stylistic markers, we produce two-dimensional feature vector, where the two components correspond to the mutual information score in either direction. 

\item {\em Aggregated Coordination Features}: For each pair, we aggregate the {\em Coordination Features} for all eight stylistic markers in both  directions, which results  in a sixteen-dimensional feature vector. 

\item {\em Length Coordination Features}: For each pair, we calculate the Pearson correlation coefficients between length of utterances in either direction, and use those coefficients as an input to SVM.   
\end{itemize}

For the Wikipedia data, we consider (admin, non-admin) pairs and for the Supreme Court data, we consider (justice, lawyer) pairs.

\setcounter{figure}{0} 
\begin{figure}[htbp]
\centering
\subfigure[]{
    \includegraphics[width=0.45\columnwidth]{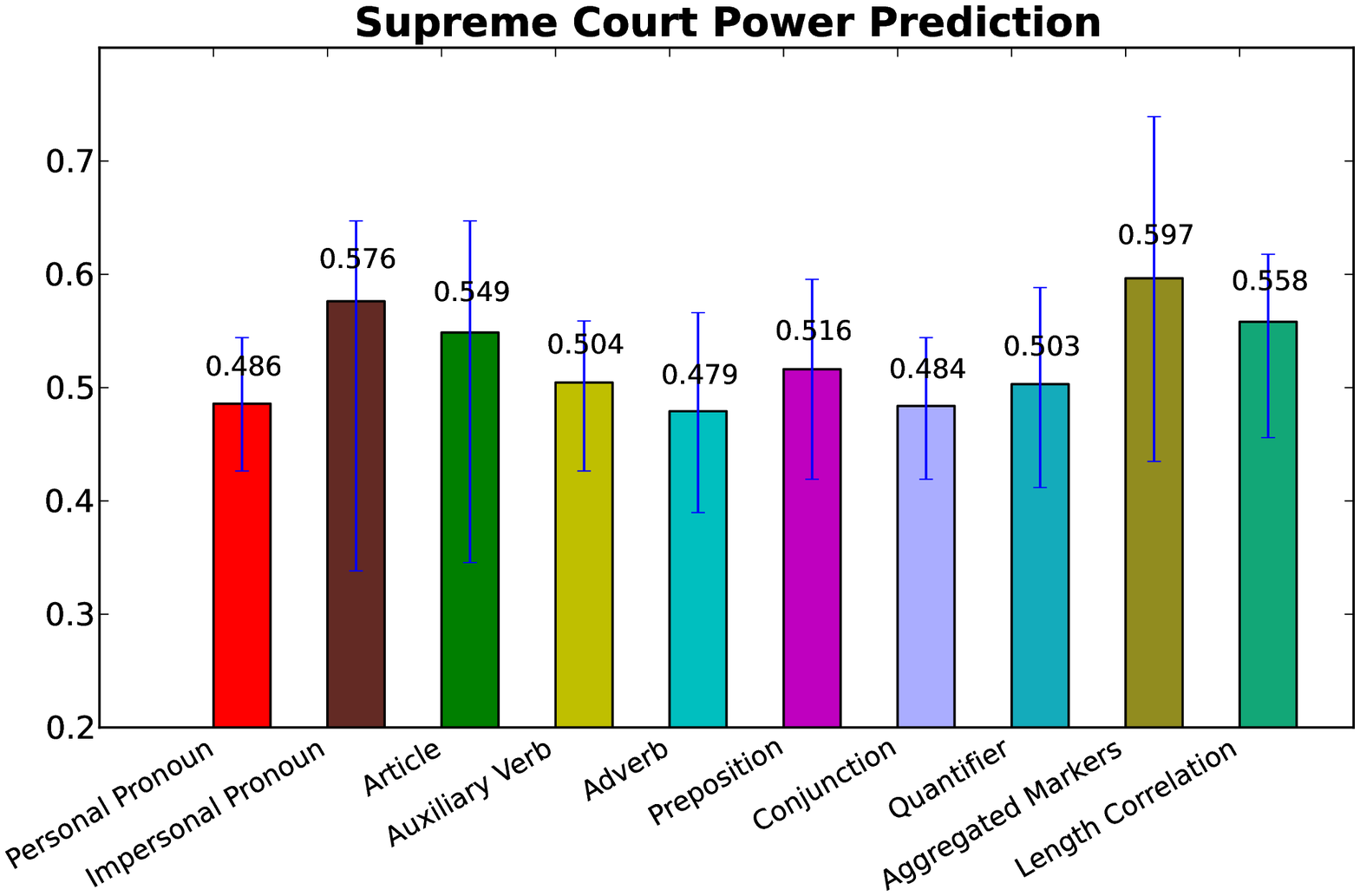} \label{fig_sc_svm}
 
    } 
    \subfigure[]{
    \includegraphics[width=0.45\columnwidth]{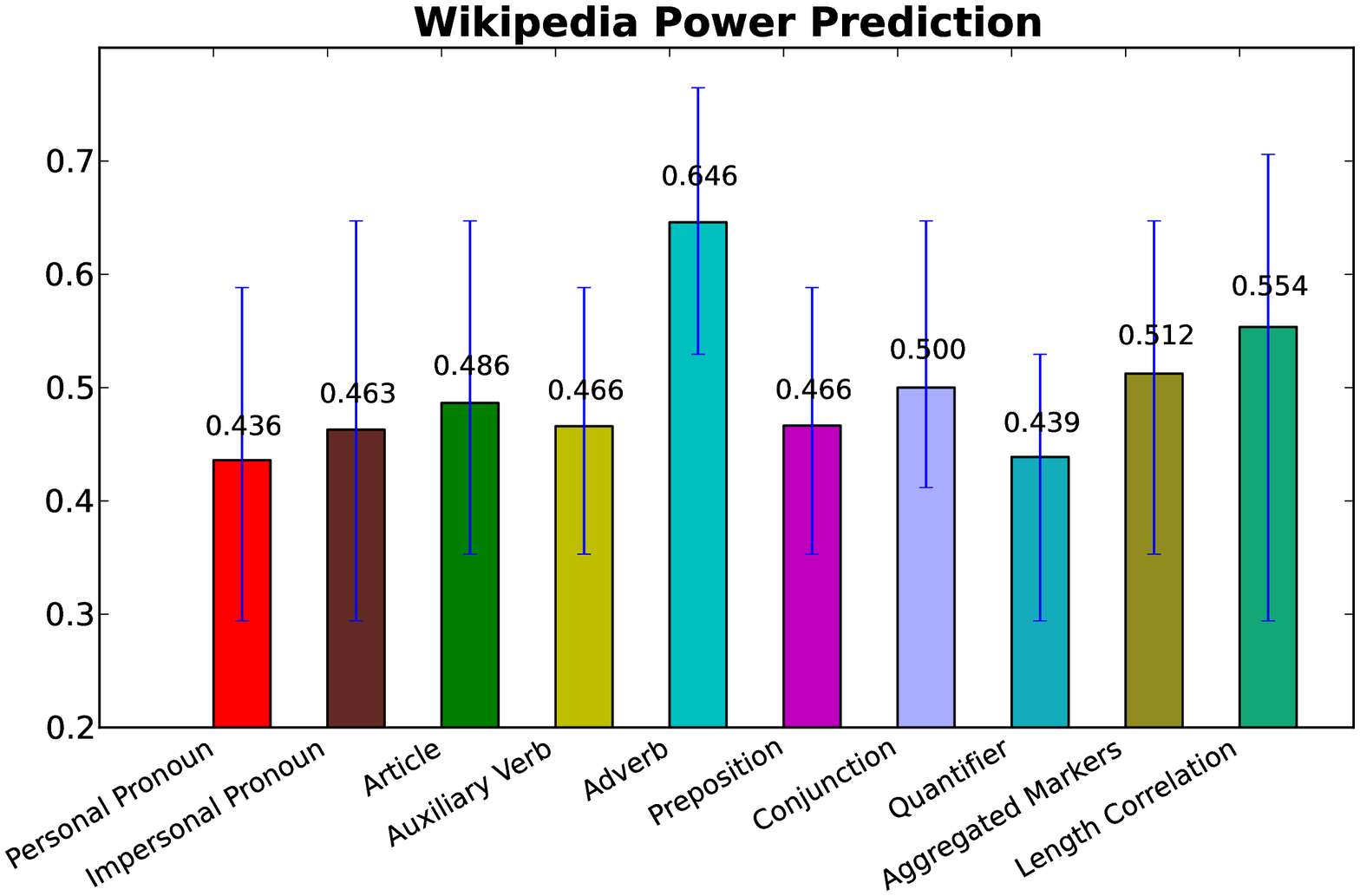} \label{fig_wiki_svm}
    }
 
\caption{SVM Prediction Accuracy for both stylistic coordination features and length coordination features }       
\label{fig_svm}
\end{figure}
In our experiment, we only select pairs which have at least 20 exchanges between them, so that we can calculate mutual information with reasonable accuracy. This results in 135 pairs in Supreme Court dataset and 34 pairs in Wikipedia dataset. Also, we labeled half fraction of the pairs, shuffled the training data and repeated the procedure $N=100$ times to calculate the average prediction accuracy. 

Fig.~\ref{fig_svm} depicts the prediction accuracy for each of the above scenario, together with error bars which give the 95\% confidence intervals. Since the two datasets are small, the error bars are relatively large in these situations. The results can be summarized as follows. For the Supreme Court dataset, the best prediction accuracy is achieved when using {\em Aggregated Coordination Features}, whereas for the Wikipedia dataset the best accuracy corresponds to using coordination on the feature  {\em Adverb}. More importantly, we find that using {\em Length Coordination Features} alone can predict user status with better-than-random ($50\%$) accuracy. In fact, investigating from the error bars, we cannot rule out that the hypothesized correlation between social status and (the direction of) stylistic coordination is due to the confounding effect of length coordination. 



\end{document}